\documentclass[journal]{IEEEtran}
\usepackage[utf8]{inputenc}
\usepackage[T1]{fontenc}
\usepackage{mathptmx}
\usepackage{amsmath,amssymb}
\usepackage{graphicx}
\usepackage{booktabs}
\usepackage{array}
\usepackage{calc}
\providecommand{\tightlist}{\setlength{\itemsep}{0pt}\setlength{\parskip}{0pt}}
\usepackage[hidelinks]{hyperref}
\usepackage{xurl}
\usepackage[htt]{hyphenat}
\begin{document}
\title{Sampling Luck Masquerades as Allocation Gain:\\ Auditing Test-Time Budget Allocation for Neural\\ Combinatorial Optimization}
\author{Jinhyung~Bae%
\thanks{J. Bae is with the Department of Industrial and Management Engineering and the AI Data Convergence program, Hankuk University of Foreign Studies, Yongin, Republic of Korea (e-mail: kevinbae2006@hufs.ac.kr).}}
\markboth{Preprint v2 (correction notice added) --- November 2026}{Bae: Auditing Test-Time Budget Allocation for NCO}
\maketitle
\begin{abstract}

Neural combinatorial optimization (NCO) solvers report the best of many sampled solutions per instance, and the number of samples is, by convention, identical for every instance. Whether a non-uniform allocation of a fixed total sample budget would buy anything has not been measured. We measure it, and we audit the measurement procedure itself.

Two findings. First, on in-distribution workloads the allocation headroom is not detectable: across three pretrained solvers (POMO, AM, SymNCO) on uniform TSP-100, an oracle allocation computed and evaluated on the same stored samples reports a gain of 2.2--2.6\% with confidence intervals excluding zero, but the same gain measured out of sample is indistinguishable from zero (0.457, 0.015 and $-$0.512 percent, all with intervals covering zero). Following the customary in-sample procedure, all three solvers would have supported a published claim of a 2\%-level allocation gain that does not exist. We quantify this bias with an instance-wise null in which the true gain is zero by construction, and show that, over the ranges we test, it does not shrink as the number of stored samples per instance or the number of instances grows: within those ranges it cannot be outrun by collecting more data.

Second, the same correction that removes the phantom gains preserves a real one. Under distribution shift --- a workload mixing uniform and clustered instances, on which these checkpoints were not trained --- a pre-registered confirmatory experiment finds that allocation guided by held-out sample statistics improves best-of-\emph{k} by 11.5\% (AM, primary endpoint; 95\% CI {[}7.4, 19.7{]}) and 12.0\% (SymNCO, replication) \textbf{at equal evaluation budget, with the cost of acquiring the guiding signal not charged}; a pre-registered negative control (POMO, whose failure under shift is an order of magnitude smaller) shows $-$0.3\% {[}$-$0.7, 0.24{]}. The gain exceeds a frozen distribution-label baseline by 4.2 percentage points {[}1.9, 7.7{]}, so it is not merely a proxy for the label. The ordering of the effect across the three solvers matches the ordering of their distribution-shift failure.

Because the registered endpoint does not charge for the guiding signal, we also report an exploratory budget-accounted policy that is executable at deployment time: a 20-sample probe charged against the same total budget, with the allocation driven by the probe's coefficient of variation and no reference tour. It retains 3.4\% (AM) and 4.6\% (SymNCO) at the registered 50:50 composition while the negative control stays at $-$0.4\%, and an exploratory sweep over workload composition suggests it retains substantially more --- 11.0\% at a 10\% out-of-distribution share --- with the relationship non-monotone in that share.

We give a correction procedure and a reporting checklist, and we release all cost arrays, analysis code, and the pre-registration record including every amendment and its direction.
\end{abstract}

\noindent\fbox{\parbox{\dimexpr\linewidth-2\fboxsep-2\fboxrule\relax}{\small
\textbf{Correction notice (v2, November 2026).} The in-distribution conclusion of this paper is wrong. The headroom curves in v1 were estimated by averaging random permutations of the stored samples (Monte Carlo curves). Recomputed with exact best-of-$k$ curves, a U-statistic over all orderings of each stored pool (uniform TSP-100, $N=50$, $S/N=100$, pool of 800), the true in-distribution headroom is 1.57\% (POMO), 1.12\% (AM) and 1.19\% (SymNCO), not zero.
\begin{itemize}\setlength{\itemsep}{0pt}
\item The v1 in-sample estimates (2.23\%, 2.57\%, 2.21\%) overstated the headroom by 0.7--1.4 points, because a greedy allocator exploits the noise in Monte Carlo curves.
\item The v1 sample-split estimates (0.49\%, 0.17\%, 0.22\%) understated it, because curves built from half of a pool are not curves of the full pool. The statement that the in-distribution gain is indistinguishable from zero, and claims that treat the sample-split estimate as the true value, therefore do not hold.
\item The out-of-distribution result stands. With exact curves the gain is 14.1\% (AM) and 12.7\% (SymNCO) in sample, and 13.9\% and 12.0\% with sample splitting.
\end{itemize}
The recomputation is reported in Appendix~B of J.~Bae, ``Give Up to Get More: Priors, Not Policies, Govern Test-Time Compute Allocation,'' arXiv:2611.06372. The text below is unchanged from v1.}}
\vspace{0.6em}
\section{Introduction}

\subsection{The convention}

Constructive neural solvers for routing problems do not emit one solution. They emit many and keep the best. The Attention Model \cite{kool} samples 1,280 tours per instance; POMO \cite{kwon} rolls out one greedy trajectory per starting node and multiplies by eight dihedral augmentations, giving 800 deterministic trajectories for a 100-node instance. In both cases the count is a single global hyperparameter. Every instance in the benchmark receives the same number of attempts.

This is a resource allocation decision made by default rather than by analysis. Instances are not equally responsive to additional samples: for some the best-of-\emph{k} curve flattens after a handful of rollouts, for others it keeps descending. If the marginal value of a sample differs across instances, then under a fixed total budget the uniform allocation is leaving something on the table. How much is an empirical question that, to our knowledge, has not been answered.

\subsection{Two questions}

\textbf{Q1 (value).} Under a fixed total sample budget, how much does instance-wise allocation buy relative to the uniform convention?

\textbf{Q2 (measurement).} The natural way to answer Q1 is to collect samples, estimate each instance's best-of-\emph{k} curve, compute the optimal allocation, and report the improvement. Decision and evaluation then use the same samples. Is the resulting number trustworthy?

Q2 is not a technicality. The allocation step is an explicit maximization over instances, and maximization over noisy estimates is exactly the setting in which selection bias is known to be severe --- the optimizer's curse in decision analysis \cite{smith}, inference on winners in econometrics \cite{andrews}, data-snooping in forecast evaluation \cite{white,hansen}. The contribution of this paper is not to discover that such bias exists. It is to quantify it in the NCO test-time allocation setting, where it has not been measured, to give a domain-specific correction, and to demonstrate that the correction discriminates rather than merely deflates.

\subsection{Contributions}

\begin{enumerate}
\def\labelenumi{\arabic{enumi}.}
\item
  \textbf{To our knowledge, the first measurement of allocation value for NCO test-time sampling.} On in-distribution workloads the value is not detectable. Under distribution shift it is 11--12\% at equal evaluation budget with the signal-acquisition cost not charged, and 3--5\% at that same 50:50 composition when a probe is charged against the total budget (rising to 11.0\% at a 10\% shifted share, exploratory); its magnitude across three solvers orders with the magnitude of each solver's failure under shift. The distribution-shift result is a pre-registered confirmatory experiment with a declared primary endpoint, a replication arm, and a negative control.
\item
  \textbf{To our knowledge, the first quantification of in-sample selection bias in this setting.} With an instance-wise null construction in which the true allocation gain is zero, the customary in-sample procedure manufactures gains of the same order as the effects it is used to detect. The bias is approximately invariant to the number of stored samples per instance and to the number of instances, over the ranges we test.
\item
  \textbf{A correction and a two-sided demonstration.} We report the gain out of sample and calibrate the in-sample estimate against the instance-wise null. The same procedure removes the in-distribution phantom gains and leaves the distribution-shift gain intact.
\item
  \textbf{A full pre-registration record.} Gate criteria, endpoints and amendments were fixed before the corresponding results were seen; every amendment is reported together with the direction in which it moves the verdict, including amendments that work against our own narrative.
\end{enumerate}

\subsection{What this paper does not claim}

We do not claim a new statistical phenomenon; the mechanism is classical selection bias, applied to a setting where it has not been accounted for. We do not claim that uniform allocation is optimal in distribution --- only that no gain is detectable at our measurement precision. We do not claim that the test-time compute literature uniformly fails to correct for this: Snell et al.~\cite{snell} use two-fold cross-validation for strategy selection, and we discuss the landscape in Section~VI. We do not claim a dose--response relationship between distribution-shift magnitude and allocation gain; we have three solvers and report an ordering.

\section{Problem Setup}

\subsection{Formulation}

Let \texttt{f\_i(k)} be the expected cost of the best of \texttt{k} solutions drawn for instance \texttt{i}. For a workload of \texttt{N} instances and total budget \texttt{S}:

\begin{align*}
\text{minimize}\quad & \textstyle\sum_i f_i(k_i)\\
\text{subject to}\quad & \textstyle\sum_i k_i = S,\; k_i \geq 1,\; k_i \in \mathbb{Z}
\end{align*}

\texttt{f\_i} is non-increasing and convex in \texttt{k}, being the expectation of a minimum order statistic: the hundredth sample cannot help as much as the tenth. The problem is therefore a convex separable resource allocation, for which greedy marginal allocation --- repeatedly giving the next sample to the instance with the largest \texttt{f\_i(k\_i)\ $-$\ f\_i(k\_i+1)} --- is optimal.

The consequence matters for how this paper is organized: \textbf{if \texttt{f} were known, the optimization would be solved. The difficulty is estimating \texttt{f} cheaply, and the difficulty of the difficulty is knowing whether the estimate can be trusted.}

\subsection{The unit of allocation}

Three quantities can be increased at inference time, and an allocation study is ill-posed until one is designated as the unit.

\textbf{Axis A --- stochastic decoding rollouts.} Temperature sampling produces \texttt{k\_i} solutions. Multi-start and augmentation are disabled. This is the only axis with no upper bound: multi-start is capped by the number of nodes, and augmentation is a fixed set of eight symmetries. It is also the canonical inference mode for the Attention Model.

\textbf{Axis B --- the finite pool of the standard protocol.} POMO's standard inference is \texttt{n} multi-start greedy rollouts times eight augmentations --- for TSP-100, a deterministic pool of 800 trajectories. Randomizing the pool order and taking the first \texttt{k\_i} without replacement makes the draws exchangeable, so \texttt{E{[}min{]}} remains non-increasing and convex and the greedy argument survives with the box constraint \texttt{k\_i\ $\leq$\ \textbar{}pool\_i\textbar{}}.

Both axes are reported. Axis B answers the objection that a study conducted purely in sampling mode measures a weakened inference procedure; Axis A answers the objection that a bounded pool cannot express arbitrary allocations. Where the two disagree, we say so.

\subsection{The allocating agent is a single solver}

The budget being divided belongs to \textbf{one fixed solver}. If different instances were served by different checkpoints, \texttt{$\Sigma$\_i\ f\_i(k\_i)} would no longer describe allocation across a workload but selection among solvers, which is a different problem with its own literature \cite{gao}. We therefore hold the checkpoint fixed across the workload.

The workload, by contrast, is unconstrained. From the perspective of a deployed solver, an out-of-distribution instance is simply a hard instance, and its difficulty is a legitimate target for allocation regardless of whether that difficulty originates in the instance geometry or in the solver's training distribution. We therefore describe our findings in terms of \emph{workload heterogeneity as seen by a fixed solver}, not in terms of intrinsic instance hardness.

\subsection{Offline replay}

For every instance we store the costs of all \texttt{K} sampled solutions. Any allocation policy can then be evaluated by taking, for each instance, the minimum of the first \texttt{k\_i} entries of a randomized ordering of its stored array. No further inference is required.

This is not a workaround for limited compute. It forces every policy --- uniform, oracle, label-based, out-of-sample --- to be compared on literally the same samples, which removes run-to-run noise from all policy comparisons. It is also what makes the audit in Section~III affordable: the null distributions are computed by resampling the stored arrays.

Costs are normalized to gap-to-reference, \texttt{gap\_i(k)\ =\ 100\ $\cdot$\ (cost\_i(k)\ /\ ref\_i\ $-$\ 1)}, with \texttt{ref\_i} an LKH-3 tour. A deterministic denominator is essential here: a denominator estimated from the same samples (for example the expected cost at \texttt{k\ =\ 1}) would be correlated with the dispersion that drives the allocation effect, inflating the measured gain as a normalization artifact.

\subsection{Estimands}

\texttt{d\ =\ (uniform\ $-$\ oracle)\ /\ uniform}, the relative improvement in mean gap. We report it through two estimators:

\begin{itemize}
\tightlist
\item
  \textbf{\texttt{d\_in}} --- allocation decided and evaluated on the same stored array.
\item
  \textbf{\texttt{d\_split}} --- the stored array is split in half; the allocation is decided from one half and evaluated on the other.
\end{itemize}

In expectation \texttt{d\_in} is biased upward, because the allocation exploits the realized noise of the array it is scored on, and \texttt{d\_split} is biased downward, because the allocation is decided from half as much data as is available. \textbf{These are bias directions in expectation, not per-realization guarantees}, and we do not present \texttt{{[}d\_split,\ d\_in{]}} as a hard interval.

\texttt{d\_split} is an out-of-sample estimate of allocation value. It is \textbf{not} a budget-accounted policy: the samples used to decide the allocation are not charged against \texttt{S}. See Section~VII (Limitation 4).

\section{The Audit}

\subsection{Regularizing the oracle}

The greedy rule requires non-increasing marginal gains. Estimated curves violate this --- in synthetic checks only 58\% of consecutive estimated marginals were non-increasing --- and the violation is not innocuous: sorting all marginals and taking the top \texttt{S\ $-$\ N} can then select the \texttt{j}-th marginal of an instance without its \texttt{(j$-$1)}-th, so the resulting counts no longer correspond to a prefix and the allocation is not a feasible policy. We therefore take the greatest convex minorant of each estimated curve before differencing. The true \texttt{f\_i} is convex, so this is regularization toward a known property, not a modelling choice.

\subsection{The instance-wise null}

To calibrate \texttt{d\_in} we need data on which the true allocation gain is zero. We construct it per instance: resample with replacement from a single instance's stored array to create \texttt{N} synthetic instances that share that instance's marginal cost distribution but are exchangeable, so no allocation can help in expectation. The synthetic set is passed through the identical pipeline, including convex regularization. Repeating this for many source instances yields a distribution of per-source 95th percentiles.

\textbf{Definition of the floor.} We report the \textbf{median} of the per-source p95 values as the primary floor and the \textbf{p90} as a conservative variant. We do \emph{not} use the maximum. A sweep over the number of source instances shows the maximum diverges: for SymNCO on Axis A it grows from 20.1 (8 sources) to 27.7 (12), 43.9 (25) and 90.4 (50), because the per-source distribution is heavy-tailed. A statistic whose value depends on how many sources one happens to examine cannot serve as a decision threshold. The median and p90 are stable under the same sweep.

An earlier version of this work used a pooled null --- pooling all instances' costs and resampling from the pool --- which we discarded: under a size- or distribution-mixed workload the pool has greater dispersion than any real instance, so the resulting floor is inflated. Numbers computed under the pooled null are not reported anywhere in this paper.

\begin{figure}[!t]\centering
\includegraphics[width=\columnwidth]{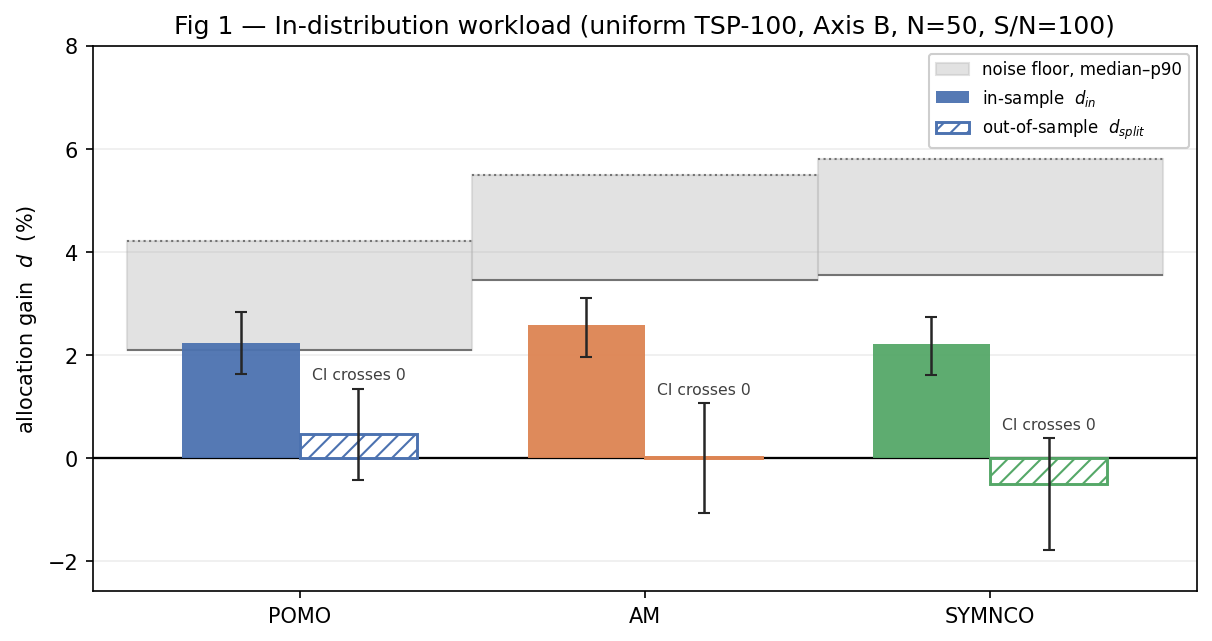}
\caption{In-distribution audit (uniform TSP-100, Axis~B, $N{=}50$, $S/N{=}100$). Solid bars: in-sample $d_{\mathrm{in}}$, all three 95\% CIs excluding zero. Hatched bars: the same gain out of sample ($d_{\mathrm{split}}$); every interval covers zero. Grey bands: the instance-wise noise floor, median to p90. Read in-sample, each solver supports a 2\%-level published gain; read out of sample, none does.}\label{fig:audit}\end{figure}
\subsection{In-distribution result}

Homogeneous workload (50 uniform TSP-100 instances), Axis B (the standard protocol), \texttt{S/N\ =\ 100}, \texttt{K\ =\ 800} pool, 95\% intervals from instance bootstrap (Table~\ref{tab:audit}):

\begin{table}[!t]\centering\scriptsize\setlength{\tabcolsep}{3pt}
\caption{In-distribution audit: uniform TSP-100, Axis B, $N{=}50$, $S/N{=}100$.}\label{tab:audit}
\begin{tabular}{llll}\toprule
Solver  & $d_{\mathrm{in}}$ {[}95\% CI{]}  & $d_{\mathrm{split}}$ {[}95\% CI{]}  & floor (med/p90)  \\
\midrule
POMO & \textbf{2.227} {[}1.63, 2.82{]} & 0.457 {[}$-$0.44, 1.34{]} & 2.083 / 4.211 \\
AM & \textbf{2.567} {[}1.96, 3.10{]} & 0.015 {[}$-$1.08, 1.06{]} & 3.440 / 5.488 \\
SymNCO & \textbf{2.206} {[}1.61, 2.72{]} & $-$0.512 {[}$-$1.79, 0.38{]} & 3.553 / 5.793 \\
\bottomrule
\end{tabular}\end{table}

Read the first numeric column alone and every row is a finding: a 2\%-level allocation gain with an interval excluding zero, on three independent pretrained solvers. Read the second column and there is nothing: all three out-of-sample estimates cover zero, one of them negative and one indistinguishable from zero at the third decimal, which is what convexity predicts when instances are exchangeable and an allocation is driven by noise.

Against the floor, AM and SymNCO sit below both the median and the p90 variant. POMO's \texttt{d\_in} of 2.227 sits marginally above its median floor of 2.083 and below its p90 floor of 4.211. We report all three quantities rather than choosing the definition that makes the row cleanest; the discriminating evidence in this paper is the out-of-sample column, which is unaffected by the floor definition.

A synthetic check with true \texttt{d\ =\ 0} by construction produces in-sample gains of 1.8--3.6\%, the same order as the table above and as the 2\% threshold we had pre-registered as a ``proceed'' criterion.

\begin{figure*}[!t]\centering
\includegraphics[width=0.85\textwidth]{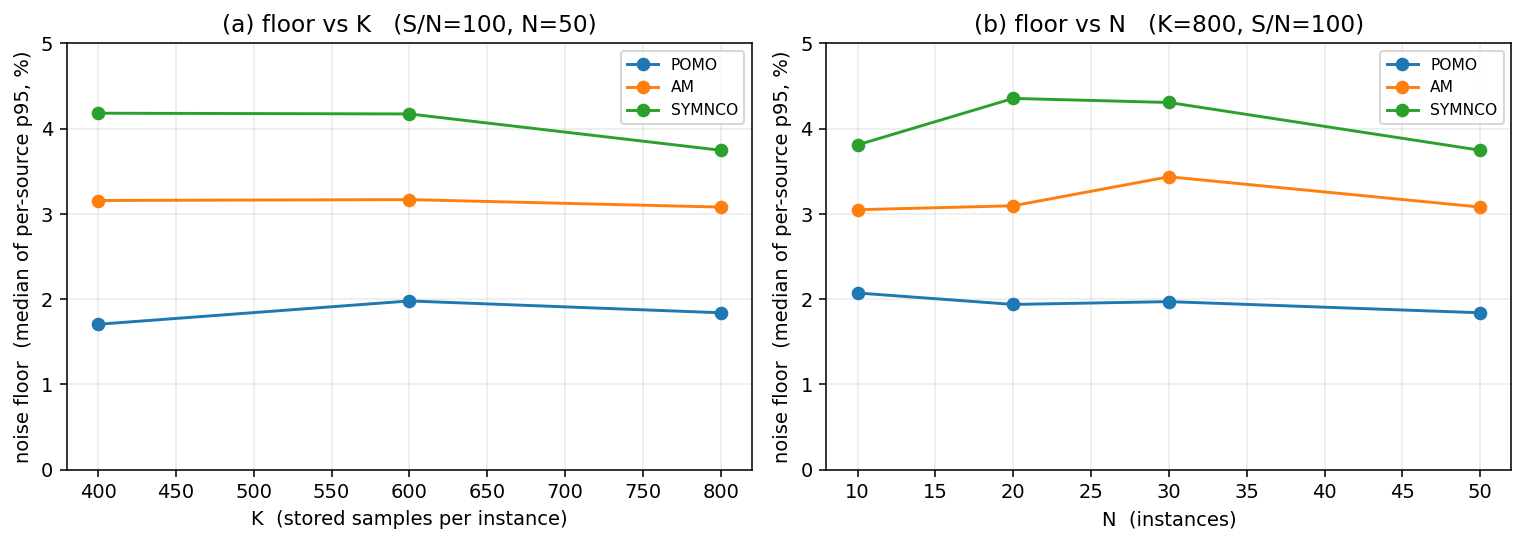}
\caption{Operating characteristics of the floor (median definition). (a) Flat in the number of stored samples per instance $K$ at fixed $S/N$; (b) flat in the number of instances $N$. Over the ranges we test, the floor does not decrease with scale; correction is the only exit we have found.}\label{fig:floor}\end{figure*}
\subsection{Operating characteristics of the floor (Fig.~\ref{fig:floor})}

The natural response to a noise floor is to collect more data. It does not work.

\textbf{(a) More samples per instance.} Axis B, homogeneous, \texttt{S/N\ =\ 100}, \texttt{N\ =\ 50}, floor (median), Table~\ref{tab:floorK}:

\begin{table}[!t]\centering\scriptsize\setlength{\tabcolsep}{3pt}
\caption{Noise floor (median of per-source p95) against stored samples per instance $K$. Axis B, homogeneous workload, $S/N{=}100$, $N{=}50$.}\label{tab:floorK}
\begin{tabular}{llll}\toprule
Solver & K = 400 & K = 600 & K = 800 \\
\midrule
POMO & 1.705 & 1.979 & 1.840 \\
AM & 3.157 & 3.168 & 3.080 \\
SymNCO & 4.182 & 4.173 & 3.747 \\
\bottomrule
\end{tabular}\end{table}

\textbf{(b) More instances.} Axis B, homogeneous, \texttt{K\ =\ 800}, \texttt{S/N\ =\ 100}, floor (median), Table~\ref{tab:floorN}:

\begin{table}[!t]\centering\scriptsize\setlength{\tabcolsep}{4pt}
\caption{Noise floor (median of per-source p95) against the number of instances $N$. Axis B, homogeneous workload, $K{=}800$, $S/N{=}100$.}\label{tab:floorN}
\begin{tabular}{lllll}\toprule
Solver & N = 10 & N = 20 & N = 30 & N = 50 \\
\midrule
POMO & 2.072 & 1.938 & 1.971 & 1.840 \\
AM & 3.049 & 3.096 & 3.436 & 3.080 \\
SymNCO & 3.810 & 4.355 & 4.307 & 3.747 \\
\bottomrule
\end{tabular}\end{table}

Both are flat within Monte-Carlo error. \textbf{The floor does not come down with more samples per instance, and it does not come down with more instances: over the ranges we test it does not come down with scale, and correction is the only exit we have found.} That is the caption of Fig.~\ref{fig:floor} and the operational content of Section~V.

Two consequences follow. Reporting a bigger experiment does not make an in-sample allocation gain more trustworthy. And a practitioner cannot infer the floor from ours --- it varies by a factor of two across solvers here (1.8 to 3.7 at \texttt{K\ =\ 800}) --- so it must be computed for the solver and configuration at hand.

\emph{Note on an earlier claim.} A previous analysis of ours reported that the floor grows with \texttt{N}. That result was produced under the discarded maximum-based definition, in which the statistic grows mechanically with the number of sources examined, and the comparison additionally confounded \texttt{N} with the number of source instances. Under the corrected definition the effect is absent. We record the retraction because it removes a result that would have supported an appealing extreme-value narrative, and because it is the second of two occasions on which our own re-measurement destroyed one of our findings (the first being the pooled null of Section~III-B).

\section{When Allocation Is Real}

\subsection{The workload}

The audit above uses an in-distribution workload, which is the condition least favourable to allocation: instances drawn i.i.d. from the training distribution are close to exchangeable, and under an exchangeable workload convexity favours uniform allocation. The interesting question is what happens when the deployed workload departs from the training distribution --- the ordinary condition in deployment.

We construct a mixed workload of 50 instances: 25 uniform TSP-100 and 25 clustered TSP-100 (four Gaussian clusters, $\sigma$ = 0.06), served by the same single checkpoint. Clustered instances are out of distribution for all three checkpoints, and the severity differs sharply between them.

\subsection{Pre-registered confirmatory experiment}

An exploratory pass over three solvers and two axes produced 12 cells, of which two were positive. Rather than report those, we pre-registered a confirmatory experiment: \textbf{new instance seed and new decoding seed, everything else frozen} (N = 100 instances, K = 1,000, \texttt{S/N\ =\ 100}, set composition, analysis code, the label baseline ratios), with endpoints declared in advance and no tests beyond them.

\begin{table*}[!t]\centering\small\setlength{\tabcolsep}{6pt}
\caption{Pre-registered confirmatory experiment: new instance and decoding seeds, all else frozen, no tests beyond the registered endpoints.}\label{tab:confirm}
\begin{tabular}{lllll}\toprule
Role  & Solver, axis, set  & $d_{\mathrm{split}}$ {[}95\% CI{]}  & Endpoint  & Result \\
\midrule
\textbf{Primary} & AM, Axis A, mixed & \textbf{11.549} {[}7.401, 19.734{]} & CI lower $\geq$ 2\% & \textbf{pass} \\
\textbf{Secondary} & AM, residual over label baseline & \textbf{4.204} {[}1.862, 7.702{]} & \textgreater{} 0 & \textbf{pass} \\
Replication & SymNCO, Axis A, mixed & \textbf{12.017} {[}5.183, 19.982{]} & CI lower $\geq$ 2\% & \textbf{pass} \\
Negative control & POMO, Axis A, mixed & \textbf{$-$0.290} {[}$-$0.720, 0.238{]} & interval covers 0 & \textbf{as predicted} \\
\bottomrule
\end{tabular}\end{table*}

The primary arm is AM on Axis A because sampling is AM's canonical inference mode. SymNCO's canonical mode is multi-start, so its Axis A arm is labelled a mechanism replication rather than a replication of practice.

The 2\% threshold is the same one used as the ``proceed'' criterion throughout the project; we did not lower the bar for the confirmatory run.

\textbf{Statement of the headline number.} Throughout, the 11--12\% figure means: \emph{allocation guided by held-out sample statistics improves best-of-k by 11--12\% at equal evaluation budget, with the signal-acquisition cost not charged} (Section~VII, Limitation 4; budget-accounted variant in Section~IV-F). We do not state the number without that qualification.

\begin{figure*}[!t]\centering
\includegraphics[width=0.85\textwidth]{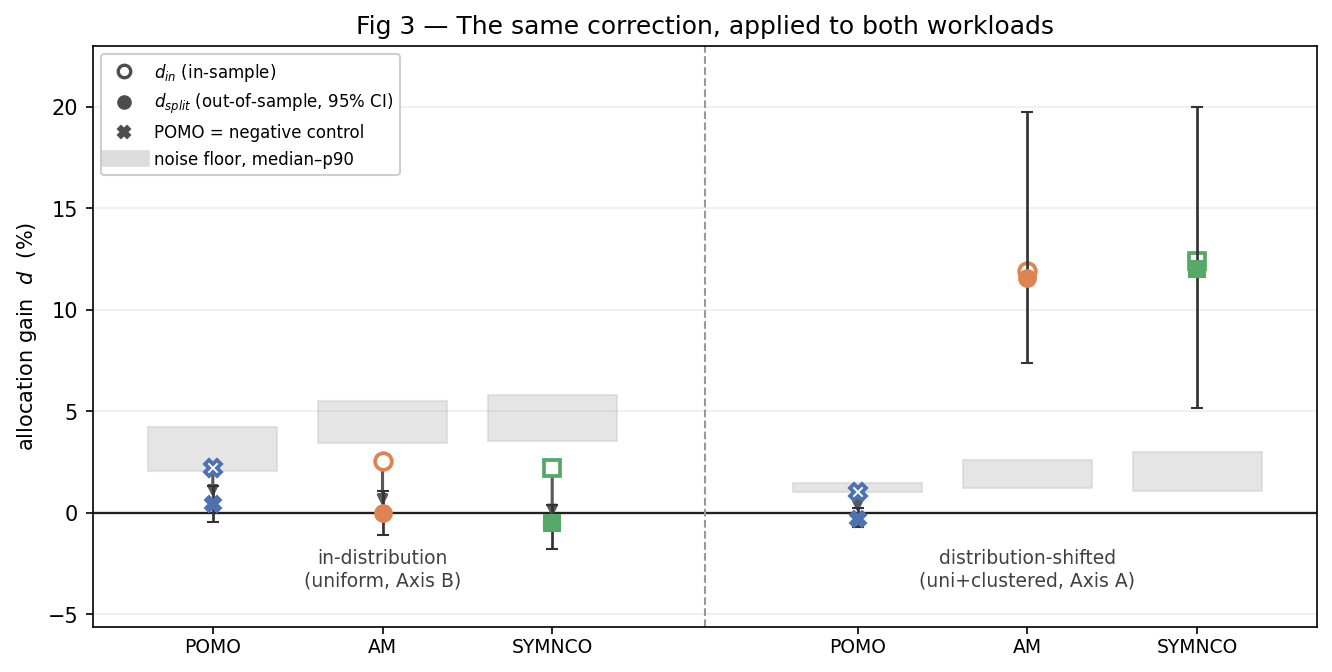}
\caption{The same correction, applied to both workloads. Left (in-distribution): in-sample estimates (open markers) sit at the 2\% level while out-of-sample estimates (filled) collapse to zero inside the noise floor. Right (distribution-shifted): the out-of-sample gain survives at 11--12\% for the two shift-fragile solvers and vanishes for the shift-robust negative control (POMO, $\times$). One procedure removes the phantom and keeps the real effect.}\label{fig:twosided}\end{figure*}
\subsection{The gain tracks the failure}

\begin{table}[!t]\centering\scriptsize\setlength{\tabcolsep}{4pt}
\caption{Distribution-shift severity per solver (mean gap on uniform vs.\ clustered TSP-100) and the corresponding out-of-sample allocation gain.}\label{tab:shift}
\begin{tabular}{lllll}\toprule
Solver & uniform gap & clustered gap & ratio & \texttt{d\_split} \\
\midrule
AM & 1.11\% & 38.63\% & \textbf{34.8$\times$} & 11.549 \\
SymNCO & 0.87\% & 23.79\% & \textbf{27.4$\times$} & 12.017 \\
POMO & 7.16\% & 17.36\% & \textbf{2.4$\times$} & $-$0.290 \\
\bottomrule
\end{tabular}\end{table}

AM and SymNCO are excellent in distribution and collapse under shift; POMO is mediocre everywhere and degrades mildly. The ordering of allocation gain matches the ordering of shift failure across the three solvers, and the negative control is the informative cell: on the solver that is robust to shift, the gain vanishes (Table~\ref{tab:shift}).

We state this as an ordering across three solvers. With three points and a gap between 2.4$\times$ and 27.4$\times$, we do not characterize the relationship as a dose--response.

\subsection{The signal is more than the label}

If the gain were simply ``spend more on clustered instances'', it would be captured by a rule that reads the distribution label and allocates in a fixed ratio. We freeze such a baseline from the exploratory pass, refit nothing on the confirmatory data, and evaluate it on the same held-out half with the same budget, distributing rounding remainders so that the label policy spends the budget exactly. The pre-registered ratios are AM 20:1 and SymNCO 12:1, the two arms in which the residual is an endpoint; the ratio applied to the negative control (POMO 1.5:1) was taken from the same exploratory pass but was not part of the registered freeze, and no endpoint depends on it. The residual --- greedy minus label, both out of sample --- is 4.204 points {[}1.862, 7.702{]} for AM.

So roughly two thirds of the AM gain is available from the label alone, and a third requires per-instance information. Both halves of that sentence matter: the trivial rule is strong, and it is not sufficient.

\subsection{Allocation does not require detecting the shift}

The obvious alternative response to distribution shift is to fix the model --- fine-tune on clustered data, or select a matched checkpoint. We do not argue against it, and the two responses are not exclusive. But they differ on two axes.

\emph{Cost.} Retraining consumes GPU time, engineering time, and labelled or generated data for the new regime. Allocation redistributes an inference budget that is already being spent; its marginal cost is zero.

\emph{Prerequisite knowledge.} Retraining must be triggered, which requires first recognizing that the workload has shifted and in what way. The allocator does not: it consumes only the solver's own sample statistics on the instance at hand, never a distribution label. This asymmetry, rather than the cost difference, is the substantive argument --- allocation is available in the interval between a shift occurring and its being detected. It is, however, not free: the sample statistics must themselves be bought, and Section~IV-F prices them.

\subsection{Charging for the signal (exploratory)}

The endpoints above do not charge for the samples used to decide the allocation. To price them we run a descriptive variant, outside the pre-registration and with no test reported, in which a fixed total budget \texttt{S\ =\ 100N} is spent by a single-pass deployable policy: draw a probe of \texttt{m\ =\ 20} rollouts per instance, \textbf{charged against the budget and retained as candidate solutions}, allocate the remaining \texttt{S\ $-$\ mN} in proportion to each probe's coefficient of variation, and report the best of each instance's \texttt{k\_i} samples. The signal uses no reference tour, so the policy is executable at deployment time (Table~\ref{tab:charged}).

\begin{table}[!t]\centering\scriptsize\setlength{\tabcolsep}{3pt}
\caption{Budget-charged probe policy (exploratory; no test reported) against the uncharged out-of-sample estimator.}\label{tab:charged}
\begin{tabular}{lll}\toprule
Solver & \texttt{d\_charged} (exploratory) & uncharged \texttt{d\_split} \\
\midrule
AM & 3.394 {[}0.609, 7.564{]} & 11.549 \\
SymNCO & 4.588 {[}1.623, 10.541{]} & 12.017 \\
POMO (negative control) & $-$0.391 {[}$-$1.002, 0.126{]} & $-$0.290 \\
\bottomrule
\end{tabular}\end{table}

Probe size sensitivity for AM: 3.36 (\texttt{m\ =\ 5}), 3.76 (\texttt{m\ =\ 10}), 3.39 (\texttt{m\ =\ 20}), 2.81 (\texttt{m\ =\ 40}).

Three observations. The effect survives budget accounting but shrinks by roughly a factor of three. The negative control continues to hold, so the discrimination between shift-fragile and shift-robust solvers is not an artefact of the free signal. And the gap between 3--5\% charged and 11--12\% uncharged is itself a result: it bounds how much a better signal than a probe's coefficient of variation could recover, and it is the natural target for the follow-up that Section~VII leaves open.

\begin{figure}[!t]\centering
\includegraphics[width=\columnwidth]{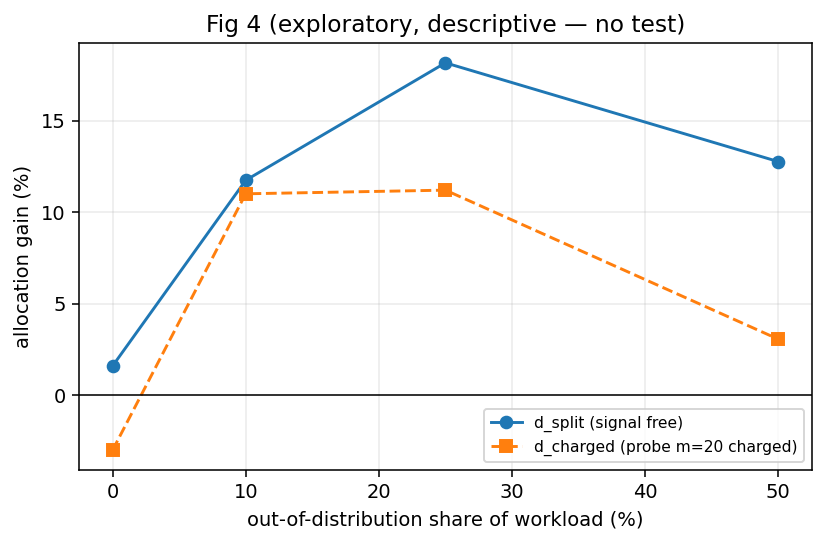}
\caption{Exploratory composition sweep (descriptive; no test): allocation gain against the out-of-distribution share, signal-free estimator vs.\ budget-charged probe policy (AM, Axis~A; $N{=}50$ subsampled workloads, 12 draws). Dependence is real but not proportional; the registered 50:50 composition is not the most favourable one.}\label{fig:ood}\end{figure}
\subsection{Gain against out-of-distribution share (exploratory)}

Subsampling the stored arrays to vary the clustered fraction, holding \texttt{N\ =\ 50} and \texttt{S/N\ =\ 100} and averaging over 12 resampled workloads. Both columns are \textbf{AM on Axis A}: the pair is \emph{the same solver under two policies} --- the signal-free estimator of Section~II-E and the budget-charged policy of Section~IV-F --- not two solvers. Descriptive only; no test is reported (Table~\ref{tab:ood}).

\begin{table}[!t]\centering\scriptsize\setlength{\tabcolsep}{3pt}
\caption{Allocation gain against the out-of-distribution share of the workload (exploratory; descriptive, no test).}\label{tab:ood}
\begin{tabular}{lll}\toprule
OOD share  & \texttt{d\_split} (signal free)  & \texttt{d\_charged} (probe m = 20 charged)  \\
\midrule
0\% & 1.586 & $-$3.018 \\
10\% & 11.740 & 11.001 \\
25\% & 18.161 & 11.201 \\
50\% & 12.760 & 3.052 \\
\bottomrule
\end{tabular}\end{table}

The 50\% row is not identical to the corresponding numbers in Section~IV-B and Section~IV-F (11.549 and 3.394) because the sweep is built from 12 subsampled workloads of \texttt{N\ =\ 50} drawn from the confirmatory pool, whereas those sections use the full \texttt{N\ =\ 100} collection once. The two are consistent, not the same estimate.

\textbf{Dependence on composition is real but it is not proportional.} The sweep suggests an interior peak: both curves rise steeply away from a fully in-distribution workload and fall back by 50\%. We state this and no more; the sweep is descriptive and carries no test, so we do not estimate the location of the peak or offer a mechanism for it. What does follow is negative: the registered 50:50 composition is not the composition most favourable to allocation, so reading our headline number as an upper bound obtained by stacking the workload with hard instances would be incorrect.

Two further readings. At a fully in-distribution workload the charged policy is \emph{negative} ($-$3.0): it spends a probe budget on a discrimination that does not exist, which is the audit result of Section~III restated as a deployment cost. And the penalty for charging the probe is small at low OOD share (11.0 versus 11.7 at 10\%) and large at 50\% (3.1 versus 12.8) --- at realistic deployment shares, the executable policy retains most of the headroom.

\section{Prescriptions}

Stated as a checklist rather than a conclusion.

\begin{enumerate}
\def\labelenumi{\arabic{enumi}.}
\item
  \textbf{Report allocation gains out of sample.} Split the stored samples, decide the allocation on one part, evaluate on the other. This is the single change that separates the phantom gains of Section~III-C from the real gain of Section~IV-B.
\item
  \textbf{If an in-sample number must be reported, calibrate it.} Build the instance-wise null at your own \texttt{(N,\ K,\ S/N)} and report the in-sample estimate against that floor. Do not borrow ours: it differs by a factor of two across three solvers on the same problem.
\item
  \textbf{Do not expect scale to fix it.} The floor is flat in both the number of stored samples per instance and the number of instances over the ranges we test. More data does not buy your way out; correction is the only exit. A larger experiment reported in-sample is not a more trustworthy one.
\item
  \textbf{Keep the budget shallow relative to the stored depth.} Estimating \texttt{f} at budgets approaching the stored array size leaves too few effective independent blocks; in synthetic checks, holding \texttt{S/N} proportional to \texttt{K} left the bias unchanged as \texttt{K} grew, while holding \texttt{S/N} well below \texttt{K} shrank it. We use \texttt{S/N\ $\leq$\ K/4} as a hard assertion in code.
\item
  \textbf{On the allocation question itself.} For in-distribution homogeneous workloads, we find no detectable headroom: uniform allocation is an adequate default within our detection limit. For shifted workloads served by a fixed checkpoint, allocation is worth measuring, it operates without a distribution label, and it retains 3--5\% at a 50:50 composition --- and, exploratorily, 11.0\% at a 10\% shifted share --- even when the guiding probe is charged against the same total budget.
\end{enumerate}

\section{Related Work}

\textbf{Conceptual ancestry.} Selecting the maximum of noisy estimates and then reporting that maximum's estimated value is the optimizer's curse \cite{smith} and the winner's curse in post-selection inference \cite{andrews}. Our \texttt{d\_in} is that construction applied to a budget allocation over instances. We cite this literature as background rather than as a quantitative prediction for our setting: the natural extreme-value scaling of a maximum over \texttt{N} does not describe our floor, which is an average over \texttt{N} instances and is flat in \texttt{N} (Section~III-D).

\textbf{Corrective tooling.} White's Reality Check~\cite{white}, Hansen's Superior Predictive Ability test~\cite{hansen}, and the Deflated Sharpe Ratio \cite{bailey} calibrate an apparently strong selected result against the distribution of results obtainable by selection alone; SIREN \cite{xu} applies the same logic to LLM benchmark reporting. The instance-wise null of Section~III-B is that family's allocation counterpart, specialized to the fact that the noise here comes from estimating minimum order statistics from a finite stored array.

\textbf{Allocation of test-time compute.} Damani et al.~\cite{damani} allocate best-of-\emph{k} per query using a learned reward-distribution predictor; Snell et al.~\cite{snell} show that difficulty-aware allocation of test-time compute is markedly more efficient than uniform; Brown et al.~\cite{brown} measure per-problem coverage curves and their heterogeneity. These establish the mechanism in the LLM setting. Our contribution is not the mechanism but the audit, and a test of whether the mechanism survives in a domain whose objective is a continuous minimum order statistic rather than a verifier-checked Bernoulli success.

On the measurement question specifically: \emph{many test-time compute studies report adaptive-allocation gains without stating that the samples used to decide the allocation are disjoint from those used to evaluate it. We do not audit any individual study here --- verifying such a separation requires the released implementation, which we did not attempt --- and we note one published exception: Snell et al.~\cite{snell} use two-fold cross-validation for strategy selection, though difficulty binning still uses oracle information. In the NCO allocation setting, no such correction is standard.}

\textbf{Neural combinatorial optimization.} POMO \cite{kwon}, the Attention Model \cite{kool} and Sym-NCO \cite{kim} supply the three checkpoints. Gao et al.~\cite{gao} select among neural solvers per instance and leave runtime-aware selection as an open problem, adjacent to the question Q1 addresses. Critiques of NCO evaluation practice \cite{liu} motivate the reporting checklist. Recht et al.~\cite{recht} provide a related example of auditing conclusions drawn from heavily reused evaluation benchmarks.

\section{Limitations}

\begin{enumerate}
\def\labelenumi{\arabic{enumi}.}
\item
  \textbf{Three points on the shift axis.} Shift severity is 2.4$\times$, 27.4$\times$ and 34.8$\times$, with nothing between 2.4 and 27.4. We claim an ordering across three solvers, not a functional relationship.
\item
  \textbf{Workload composition is a design choice, and not the most favourable one.} The registered workload is 50\% clustered, so the result should be read as ``11--12\% at this composition''. We initially expected the gain to fall roughly in proportion to the shifted share; the exploratory sweep of Section~IV-G contradicts that expectation, and we state the correction rather than quietly dropping it. Composition dependence is real but not proportional, and the sweep suggests an interior peak; the budget-charged policy is \emph{more} favourable at low shares (11.0 at 10\% versus 3.1 at 50\%). We do not estimate where the peak lies or why. Characterizing the dependence properly requires the confirmatory treatment that Section~IV-G does not have, and until then no claim in this paper rests on it.
\item
  \textbf{A single problem class.} All results are TSP-100. CVRP was gated out before collection: the reference solver's seed-to-seed spread on CVRP-100 was 0.80\% at a 5-second limit and 0.56\% at 15 seconds, against a pre-registered stability requirement of 0.05\% --- the denominator would have been as unstable as the effect under study. Three solvers with different decoding distributions and different absolute gap levels (1.0--2.3\% in distribution) substitute only partially for problem diversity. A registered fallback exists for a future revision: a fixed-seed reference is deterministic and therefore uncorrelated with the sample arrays, and since all our claims are relative comparisons between policies on a common denominator, its inexactness cancels; it would only forfeit comparability of absolute gaps to other papers.
\item
  \textbf{The registered endpoint is not budget-accounted.} \texttt{d\_split} decides the allocation using half of each instance's stored array --- 500 samples, five times the evaluation budget of 100 --- and evaluates on the other half. Those samples are \textbf{not} charged. \texttt{d\_split} therefore estimates the \emph{headroom} available to allocation, not the performance of a realizable fixed-budget policy, and we do not describe it as a policy anywhere in the text. The exploratory variant of Section~IV-F prices the signal and retains 3--5\%; that variant is descriptive, was not pre-registered, and no test is reported on it. Closing this gap properly --- designing a signal that recovers more of the 11--12\% headroom at a charged probe cost --- is left to follow-up work.
\item
  \textbf{Transparency of amendments.} Every change, with its date and the direction in which it moves the decision criteria, is tabulated in Appendix B; we do not attempt a net tally here, because the changes are not commensurable. Two illustrate the range. Replacing the pooled null with the instance-wise null (Amendment 1) \emph{lowered} the floor and so favoured detection, and we made it anyway on grounds that hold regardless of outcome --- a pooled null is invalid under a mixed workload because it fabricates synthetic instances more dispersed than any real one. Replacing the maximum-based floor with the median, made after the gate, works against this paper's own audit narrative: it moves POMO's in-distribution row to the margin of its floor, and --- together with separating the number of null source instances from \texttt{N} --- it retracted our earlier claim that the floor grows with \texttt{N} (Section~III-D). We adopted it because the definition is correct, not because it helped.

  The decision rules themselves were never changed retroactively. One known defect is therefore recorded rather than repaired: the termination rule was keyed on \texttt{d\_in} against the floor, so it could not credit the positive out-of-sample evidence that \texttt{d\_split} supplies, and it returned ``signal absence'' in all four gate cells (Appendix B, v8). We report that verdict as registered and the \texttt{d\_split} evidence alongside it.
\end{enumerate}

\section{Conclusion}

Instance-wise allocation of a test-time sample budget buys nothing measurable for neural combinatorial optimization solvers on the workloads those solvers were trained for. On a workload half of which lies outside that distribution it buys 11--12\% at equal evaluation budget with the guiding signal free, and an exploratory policy that charges a 20-sample probe against the same budget retains 3.4\% at that composition and 11.0\% at a more realistic 10\% shifted share --- in every case without needing to know that the shift occurred. The procedure conventionally used to measure such gains --- deciding and evaluating the allocation on the same stored samples --- manufactures gains of 2.2--2.6\% on data where the true gain is zero, an amount that, over the ranges we test, diminishes neither with more samples per instance nor with more instances. Reporting the gain out of sample removes the phantom and leaves the real effect standing.

\appendices
\begin{table*}[!t]\centering\scriptsize\setlength{\tabcolsep}{5pt}
\caption{Pre-registration timeline. Every gate, endpoint, and amendment was fixed before the corresponding result was seen; each amendment records the direction in which it moves the decision criteria.}\label{tab:prereg}
\begin{tabular}{p{1.2cm}p{0.6cm}p{9.2cm}p{4.6cm}}\toprule
Date (2026)  & Ver.  & Change  & Direction / note \\
\midrule
Aug 6 & v1 & Initial preregistration: allocation-gain question; gate (progress $\geq$ 2\%, terminate \textless{} 0.5\%); all-branches-reportable design & --- \\
Aug 6 & v2 & Allocation unit fixed (stochastic rollouts); mixed/homogeneous dual sets; positioned against LLM test-time-compute literature & pre-data \\
Aug 6 & v3 & Finite-pool axis (standard N$\times$8 protocol) promoted to co-primary; gap-to-optimal normalization (LKH-3); trivial baselines (size-proportional, quantized) added; CI-based gate & pre-data \\
Aug 6 & v4 & After synthetic pipeline validation: bracket \texttt{{[}d\_split,\ d\_in{]}}, no point estimates; noise-floor requirement; termination split (signal absence vs practical negligibility, the latter qualified by floor \textless{} 0.5\%); one scale-up only; \texttt{S/N\ $\times$\ 4\ $\leq$\ K}; convex-hull regularization of marginal gains & tightened against false positives \\
Aug 12 & v5 & \textbf{Amendment 1}: pooled $\rightarrow$ per-instance null & favors detection (lowers floor); justified independently --- the pooled null is invalid under mixed sets regardless of outcome \\
Aug 12 & v5 & \textbf{Amendment 2}: restrict sizes to matched checkpoints; n = 200 $\rightarrow$ labeled OOD arm & disfavors gate (reduces heterogeneity); opposite direction to Am. 1 \\
Aug 12 & v6 & \textbf{Amendment 3}: core set n = 100, distribution contrast, single checkpoint (supersedes Am. 2's remedy) & direction mixed; purpose is problem identity (single-solver allocation), not gate advantage \\
Aug 12 & v7 & Scale-up run registered with three-way outcome rules fixed pre-run & pre-result \\
Aug 13 & v8 & Gate judgment: all four cells terminate (signal absence) under the registered rule. \textbf{Amendment 4} recorded as a design defect only --- no retroactive rule change (termination keyed to \texttt{d\_in} cannot credit positive \texttt{d\_split} evidence) & rule left intact after results \\
Aug 13 & --- & Floor definition: max $\rightarrow$ median of per-source p95 (max diverges under heavy tails) & disfavors audit narrative (one \texttt{d\_in} cell becomes borderline); adopted because the definition is correct \\
Aug 13 & --- & Confirmation preregistered: AM primary ($\geq$ 2\%), label-residual (\textgreater{} 0, frozen ratios), SymNCO replication, POMO negative control with predicted direction; new seeds; one run only $\rightarrow$ all four endpoints passed & pre-result \\
Aug 13 & --- & Pre-drafting corrections: N-scaling claim retracted (artifact of max definition + source-count confound); signal-cost non-charging disclosed; charged variant added as exploratory & claims reduced after self-remeasurement \\
\bottomrule
\end{tabular}\end{table*}
\section{Reproducibility}

\textbf{Checkpoints.} \texttt{ML4TSPBench/\{POMO,AM,SYMNCO\}} --- \texttt{tsp100\_pomo.pt}, \texttt{tsp100\_am.pt}, \texttt{tsp100\_symnco.pt}. These are legacy-format \texttt{rl4co} state dicts and require key remapping to load into current \texttt{rl4co}: feed-forward blocks \texttt{module.0/2.\{weight,bias\}} $\rightarrow$ \texttt{module.lins.0/1.\{weight,bias\}}; decoder \texttt{logit\_attention.*} $\rightarrow$ \texttt{pointer.*}; the AM archive additionally contains a rollout baseline copy under \texttt{baseline.*}, which must be excluded. Encoder depth is 3 (not the library default) and normalization is batch; both are inferred from the checkpoint. Loading asserts zero missing and zero unexpected keys --- a partial load would silently mix randomly initialized layers into the measured curves.

Verification against LKH-3 on 8 held-out instances: POMO multi-start greedy gap 2.27\%, SymNCO 1.21\%, AM sampling(512) 1.01\%.

\textbf{Collection.} N = 100 instances per run (50 uniform, 50 clustered), 100 nodes. Axis A: K = 1,000 stochastic rollouts per instance. Axis B: the full \texttt{n\ $\times$\ 8\ =\ 800} deterministic pool. Evaluation budget \texttt{S/N\ =\ 100} with \texttt{assert\ S/N\ $\times$\ 4\ $\leq$\ K} and a per-instance cap \texttt{k\_i\ $\leq$\ K/2}. References from LKH-3 (\texttt{elkai}). Arrays are stored per solver and per axis with SHA-256 digests; the confirmatory run replicates the collection code with only the instance and decoding seeds changed.

\textbf{Seeds.} Instance generation, decoding, curve ordering and bootstrap are separated and reported. Exploratory run: instance 20260812, decoding 11. Confirmatory run: instance 20260813, decoding 77. Ordering (22) and bootstrap (33) frozen across both.

\textbf{Reproduction check.} Re-collecting POMO with identical seeds reproduced the stored arrays bit-for-bit (\texttt{fd5e6a6adb646854}, \texttt{9b6146156f2fa124}).

\textbf{Artifacts.} Every number in this paper is traceable to one of six files: \texttt{expand\_report.txt} (three solvers, in-distribution audit), \texttt{confirm\_report.txt} (pre-registered confirmatory endpoints), \texttt{floor\_definition\_sweep.json} (per-source null percentiles, the median/p90/max comparison of Section~III-B, and the floor values of Section~III-C), \texttt{fig2\_floor\_median.json} (floor sweeps over \texttt{K} and \texttt{N}, median and p90), \texttt{charged\_variant.json} (budget-charged policy and its probe-size sensitivity), \texttt{fig4\_ood\_share.json} (composition sweep). Figures: \texttt{fig1\_audit\_headline.png}, \texttt{fig2\_floor\_median.png}, \texttt{fig3\_two\_sided.png}, \texttt{fig4\_ood\_share.png}. Raw cost arrays: \texttt{\{expand,\allowbreak confirm\}\_\allowbreak\{POMO,\allowbreak AM,\allowbreak SYMNCO\}\_\allowbreak axis\{A,B\}.npz}. Analysis scripts are released with the arrays; the audit, the confirmatory analysis and the two exploratory variants are separate scripts, so the pre-registered path can be re-run without the exploratory additions. All arrays, scripts, and the pre-registration record are available at \url{https://github.com/nepersoned/best-of-k-allocation}.

\section{Pre-registration Timeline}

All gates, endpoints, and amendments were fixed before the corresponding results were seen. Each amendment records the direction in which it moves the decision criteria; the full record is Table~\ref{tab:prereg}.

\end{document}